# Principle Interference in Technical and Scientific Translation

**Mohammad Ibrahim Qani**

Assistant Professor (Pohanyar), Faryab University, Afghanistan. 2023

## Abstract

In this article, I will explore the nature of interference in translation, especially in technical and scientific texts, using a descriptivist approach. I will have a brief overview of the historical excursion of interference in technical and scientific translation. My aim is to explain this phenomenon and its causes with all its paradoxes, instead of simply condemning it as an example of supposedly bad translation. Thus, I will focus on its status in the bibliography of translation, on the motives for and consequences of interference in specialized translation, as well as on the nature of the arguments given for and against this phenomenon.

Therefore the relationship between different societies has always been possible with the act of translation. When civilizations are examined throughout history, it is seen that the dissemination of knowledge among different societies has been achieved by translation. These societies have often become aware of the advancements in technology and science by means of translation. Therefore; translation becomes very significant in technical contact between societies and humans. Since the translation of technical texts is the preliminary scope of this thesis, it will be beneficial to have a brief look at the history of technical translation in the world.

## Keywords

Interference, translation, terminology, inter-language, descriptivism, prescriptivism, globalization, technical translation,

**Aim:** By studying this article, readers will get awareness of historical achievements and the main interference in scientific and technical translation. They will learn how to render words from the source language into the target language considering technical and scientific translation methods.

## Methods of Research

In this article, I used the APA style and the usual and old research method which is called Library Research (data collection), online libraries, PDF books, available research papers in reliable journals and websites, and some other sources. This article covers the analysis of rendering some sort of technical and scientific words from the source language into the target language.

## Introduction

Translation is becoming a crucial element and part of our life. Today's world extremely needs the arts of translation because the world today is getting as close as possible; people around the world have been trying to conduct better connections and relations with each other. Technical and scientific translation is one of the very important topics in the translation field. Recently, it is covered the most important place in the translation market. It is an extremely vital subject in linguistic universities and much research has been done in this area by scientist around the world. But this subject is nearly new and unprecedented in Afghanistan's universities. This research will help those university professors and lecturers who are going to translate different books from different languages in order to get an academic scientific ranking.

## A General History of Technical Translation

Since the creation of human beings, communication has always been a method used to establish contacts. Then we can say that the act of translation started together with the communication. Since communication is a must in social life, the need for sign systems and translation is unavoidable. Communication can be both within the limits of a society and between different societies that use different sign systems.

The Middle ages comprise the period between the fifth and fifteenth centuries. The twelfth century is the period when translation activities reached a peak. Lots of scientific works of eastern civilizations were introduced to the West by means of translations. On the other hand, scientific books of Greek writers were translated into Latin which was considered to be the Lingua Franca of the time. James of Venice was the first translator in that period who translated from Greek to Latin. In 1182, he translated Aristotle's Organon which was called then New Logic. Another book that was translated from Greek to Latin in that period was Ptolemaeus Almagest. Ptolemaeus was an astronomer, a mathematician, and a geographer and he dealt with the complex movements of the stars and planetary paths in his book Almagest which was translated by Eugene from Sicily. Meno and Phaedo of Plato were also translated in this period by Aristiphus from Greek into Latin. When Constantinople was invaded by Latins in 1204, lots of Greek books were carried to Europe and this increased the number of translations in the thirteenth century. Aristotle's books Rhetoric, Poetics, and Metaphysics were translated into Latin (Kelly, 2009: P. 481).

The twelfth century is known as the period in which the eastern and western worlds met through translation. Adelard an English philosopher and mathematician collected lots of works of art belonging to eastern scholars and translated these works from Arabic into Latin. Furthermore, İhsa'ül Ulum of Farabi, a Turkish scholar, and Makasid'ül Felasifi of Gazali and some works related to astronomy and philosophy were translated into Latin between 1135 and 1153.

What is worth mentioning in terms of technical translation, especially in the twelfth century is the translation school of Toledo established in Spain. Arabic, Jewish and Christian cultures were brought together in this school and lots of technical texts were translated. The aim of the scholars in Toledo school was to transfer the cultural and scientific heritage of Arabian and Greek civilizations to the Central West. According to Mine Yazıcı Toledo school had a great impact on the dissemination of scientific and philosophical knowledge to Europe Works of İbni Sina were also translated into Latin in this school. Gerard, a translator of Italian origin, is considered to be one of the hardest working translators in Toledo school. He translated 83 scientific books into Latin. Books of İbni Sina, Farabi, and İbni Heysem were translated by him. (Neubert, 1990: P. 78)

Moreover, an Italian scholar Plato translated Batlamyo's Quadripartitum and lots of other works about geometry into Latin and a Spanish doctor Marc translated the Holy Quran and works of Hippocrates and Calinos into Latin (Kelly, 2009: 481). In the thirteenth century, translations from Arabic into Latin continued. Alfred a British philosopher translated İbn-i Sina's Şifa. Arabic interpretations of Aristo's works were translated by a Spanish translator Peter Gallego. Stephen from Zaragoza translated İbnül-Cezzar's work Edviye in 1233 and an Italian translator John translated İbn-i Sina's Elercüzetu Fıttıb. Moreover, Faraj, a Jewish translator from Sicily translated Razi's Elhavi in 1279 and İbn-i Cazla's Takvi-Mülebdan in 1280. Other translations of Muslim scholars made together with these translations in that period provided the scholars in Europe to have new insights into science. Scientists throughout Europe studied these translations in the newly established universities (Faruqi, 2006: PP. 391-399).

## Interference in Translation Studies

In an attempt to provide a wide definition for interference in translation, we could say that it is the importation into the target text of lexical, syntactic, cultural, or structural items typical of a different semiotic system and unusual or non-existent in the target context, at least as original instances of communication in the target language. This definition includes the importation, whether intentional or not, of literal or modified foreign words and phrases (lexical interference), forms (syntactic interference), specific cultural items (cultural interference, proper nouns included), or genre conventions (structural or pragmatic interference).

Interference has always been a topic of great interest in the theory of translation, although considered from different perspectives and under different labels, some of them even more value-laden than "interference" itself, such as contamination, code-switching, hetero-lingualism, linguistic influence, hybridity, borrowings, interlanguage, translationese, pidginisation, anglicis action (or whatever the source language), Spanglish, Polish (or whatever the language pair), interpenetration or infiltration, just to mention a few. Lexical and syntactic interference in particular has traditionally been regarded as classic howlers, something to be systematically avoided because it worked against a fluent and transparent reading.

To start with the paradoxes involved in the notion of interference, its mere presence shows that the text is a translation, refuting the illusion of sameness through an excess of similarity. From this perspective, a translation using words or syntactic structures clearly derived from the original language cannot stand as a complete replacement of the source text; that is, a translation should be the same as the source text but should not sound as if it was the source text. Classic statements such as Cicero's or Jerome's defense of sense for sense as opposed to word for word translation may thus easily be read as a rejection of interference because it hampers fluency, transparency, and the full development of the target languages (TLs) as vehicles of culture in their own right. (Kelly, 2009: P. 48)

In August 2008, there were over 650 references in BITRA (Bibliography of Interpreting and Translation) to publications dealing specifically with interference in translation, and this figure does not take into account all the handbooks and publications where this issue is always present although it is not the central topic of the text.

A great majority of these texts have been published after 1950 when linguistics began to address contrastive issues of usage in modern languages in a systematic way. As was only to be expected, most of them were and still are mainly concerned with providing recipes to avoid interference in translation, especially when the language pair involved is historically close and there are numerous cognates e.g. romance languages.

Simultaneously, there have always been advocates of different levels of interference, usually when the sacred or canonical nature of the source text seemed to make it advisable to demand a special effort from the reader in exchange for a more conservative rendering, i.e. for a rendering of the source text on its own terms. Bible translation is a clear example of this and the reason why a defender of sense for sense translation such as Jerome says that in the Bible even the order of the words is sacred and should be respected. Schleiermacher, another theologian, is probably the first scholar to defend in a systematic way what could be termed 'controlled interference' in the translation of canonical and sacred texts for the same reasons. He was also probably the first to explicitly exclude technical texts from this kind of strategy since their wording supposedly did not convey any special national spirit and their translation was a mainly mechanical task. (Schleiermacher, 1813: P. 47)

**The modern times of interference**

In modern times, scholars such as Benjamin, Berman, and Venuti have retaken Schleiermacher's stance from different starting points and ideological agendas to favor overt translations enabling the reader to perceive the source text as portraying a different culture. These authors denounce normalization (i.e. the replacement of foreign or idiosyncratic marks included in the source text by the most usual variants according to target text conventions) as a strategy that eliminates "otherness" from a foreign text which should also convey a different worldview for TL addressees. Normalization, then, would result in target texts all written in a uniform way, giving the impression that all literature and views of life are essentially the same. Once again, all of these authors explicitly or implicitly eliminate technical translation from this equation, since these kinds of texts are somehow seen as international or culturally neutral. (Venuti, 1998: P. 355)

All these attempts to promote significant degrees of interference in at least certain types of translation clashed and still clash with the rejection of overt versions by publishers and readers, who are not generally prepared to accept translations whose structure and wording do not attempt to belie the asymmetrical nature of languages and cultures. Generally speaking, receivers do not like having to make an additional reading effort to understand and cope with texts bearing many lexical and stylistic instances that run contrariwise to what is considered to be optimized according to the conventions for that text type in the TL. From a theoretical point of view, relevance theory, represented in translation studies by Gutt (1991) describes this mode as a direct translation. A direct translation would provide the highest possible degree of resemblance to the original but would require the reader to process the target text using the context of the original, which is seen as fairly unrealistic since we all use our own context in order to understand. (Gutt, 1991: P. 122)

In the specific case of interference in technical and scientific translation, this is also clearly the case. With the possible exception of a sworn translation, where an important degree of literalness is usually expected in order to legally consider that the text is really 'the same, to my knowledge, there is virtually no publication asking for any kind of controlled interference in order to maintain the world view portrayed in the source texts. Indeed, even in the case of a sworn translation, apart from a great majority of practical texts on pedagogical and professional issues which do not address this topic, what we usually find regarding interference is calls to minimize it in order to obtain more functional or acceptable translations (Ramos, 2002: P. 158).

This centuries-old debate between advocates and opponents of interference, characterized by the defense of ways of translating according to the scholar's agenda, only began to change when translation studies became an autonomous discipline in the 1980s. The new attempt to replace impressionism with scientific methodology in the study of translation involved studying translation phenomena with a non-prescriptive approach. Thus, as early as 1978, Toury was already claiming that interference ("interlanguage") was very likely universal in translation and that confining its study to "error analysis" involved a serious case of simplification because in many instances interference was "preferred to 'pure' TL forms", and that it "should form an integral part of any systematic descriptive study of translation as an empirical phenomenon" (Toury 1978: PP. 224-225).

The main advantage of this approach is that it allows the researcher to explore reality instead of just judging it according to impressionistic standards. The aim is not to provide recipes for supposedly better translations whatever the context, but to explain them, to try to shed some light on facts. The underlying rationale is that a non-prescriptive understanding of the phenomenon will enable translators to act consciously and to decide for themselves which strategies to apply after obtaining a complete picture of all the possibilities, motives, and consequences. This, too, will be my approach here, in an attempt to begin to explain interference as forming part of translation, its causes, and the nature of the arguments for and against it in the bibliography of our discipline.

**Interference in technical and scientific translation**

Interference is at least as close as can be to a universal in translation and is still generally perceived as an error, especially in non-canonical technical and scientific texts, which are generally not thought to convey any sort of specific world view, either there must be some kind of rational, understandable range of motives for its use, or translators are simply incompetent. The latter seems a poor explanation: if this was so, publishers, proofreaders, and editors would simply look for competent professionals and take care to avoid this behavior because readers; especially technical readers at that would complain about unreadable or unacceptable translations which hampered information flow.

There are four main motives for interference in translation, which can be defined separately but tend to overlap in practice; the double tension intrinsically associated with translation, the creation and preservation of specific terminology or jargon, the non-existence of a given term or structure in TL, and the prestige of the source culture. All of these are present in all kinds of translation, but the last three are especially visible in scientific and technical translation.

Translation always operates between two forces, centripetal and centrifugal, which simultaneously and paradoxically push it towards the source-text proposals and towards the target-context notions of correction and optimal writing. The attraction exerted by the source text is a centripetal force which on its own would arise in translations full of interference, but it is compensated for by the centrifugal force derived from the conventions of the target context, which define "correction" according to the receiving context and, with very few exceptions, partly overlap those according to which the source text was written. This partial overlapping of norms and conventions also means that the border between interference and TL correction is often fuzzy. Since translators usually wish their texts both to represent the original and to be optimal texts in their own right according to the conventions accepted by their TL readers, inevitably, translations, whether technical or not, show a combination of both forces to different degrees, depending on how much the translators want or are able to make their texts to look like AN original or THE original. This first motive is present by definition in all translations having a minimum complexity and is the reason why interference can be considered akin to a universal in translation. This is also the character we first detect when pointing out that a given text looks like a translation, making it an inherent feature of our mental image of cross-lingual mediation (Tirkkonen & Condit, 2002: P. 207).

The centripetal force exerted within this double tension or attraction to the source and target contexts is also supported by a very powerful stimulus the economy of effort, which seems to make translators, who usually work under very tight deadlines and for a rather modest remuneration, tend to deviate from the source text only when they consider it really necessary since conservative translation is the fastest and most economical way of working.

To finish with the double tension motive, it is necessary to stress that the centrifugal force involved in this double drive is also always present, encouraging the translators to deviate from the source text in order to meet the (supposed) expectations of their readerships. The translators, then, are forced to constantly negotiate and navigate between two opposing stimuli, resulting in various historical, text-type, and idiosyncratic balances whose study forms a very important part of research in translation studies. (Neubert, 1990: P. 108)

**Terminology of jargon in interference with technical and scientific translation**

The creation and preservation of specific terminology or jargon are simply a characteristic inherent to mankind. Any group of persons sharing a profession or a common interest tends to create its own terminology for two main reasons - necessity and exclusivity. Regarding necessity, any human activity aims to have its own terminology in order to gain precision and clarity. You need the word ' starboard' because this is not relative, whereas ' right' is, and you simply need clarity and precision if you have to shout instructions in the middle of a storm on a boat. The quest for bi-univocity (one term per object/concept, and one object/concept per term) in technical terminology is a natural consequence of it (and the failure to achieve bi-univocity in most technical and scientific disciplines is one of the worst headaches for technical translators, but that is another story). Regarding exclusivity, the creation of a specific terminology brings about an important degree of opacity for outsiders, something that is generally enhanced by insiders, since it strengthens their feeling of belonging and sets their trade, vocation, or situation apart from all other mortal souls. This is quite easy to understand in the case of teenagers or criminals, but the same applies to any branch of knowledge, such as lawyers, doctors, or translation-studies scholars. (Hansen, 2002: P. 303)

From the point of view of the technical and scientific translator, there are many text types in which, as Toury writes in the aforementioned quotation, interference is indeed preferred to ' pure' TL instances. It is not strange, either, to witness translation students complaining about teachers who instilled in them maximum respect for TL purity, when their translations are professionally rejected for being so pure that they are hard to accept by their specialist readerships.

How often are novice translators surprised, perhaps even shocked at the reaction of subject specialists who re-translate certain passages of a nicely TL-worded text because they insist on terms and phrases that the TL-conscious translator had expressly eliminated? But the experts' notion of what satisfies a particular technical text class is a far cry from the translator's concept of a good TL instance. In other words, the impact of translation, specialist translations at that, is no longer felt as un-TL. The opposite is the case. That the SL-patterns look through is regarded as a perhaps novel, but certainly an in-feature of many modern normal TL texts, especially of a scientific or a technical nature, e.g. medicine, physics, electronics, etc. The impact of translation, in our epoch, is to a growing extent multidirectional. It is true that individual TLs each cope with this verbal influx in their own specific ways. But the actual outcome, however, varied it may be from TL to TL, is also invariably marked by much internationalism. (Neubert, 1990: PP. 98, 100)

**Prescriptivism and Descriptivism in the Choice of Terminology in Technical and Scientific Translation**

Now it seems appropriate to try to summarize the reasons usually given for and against interference in technical and scientific translation against the background of motives discussed above, which is characterized by a systematic and paradoxical double force.

As mentioned earlier, we can say there are two main approaches or methodological poles when addressing the study of interference in technical writing, which could be termed descriptivist and prescriptivist respectively. These two approaches are also applicable to the act of translating. To define them in as few words as possible and in their most extreme instances, descriptivists think that translators should adapt to their readers' usage, even if this is not very logical or may be questionable for any other reason. Prescriptivists, on the other hand, think that the most correct term from the point of view of absolute respect to TL traditional patterns should always be promoted, even if this means swimming against the tide.

Of course, the criteria for correction are the crucial issue here. They are all derived from the second pole of the double-tension dichotomy pointed out earlier, favoring the idea of the target text being like AN original. Usually, these criteria can be summarized as one of the three following or a combination thereof: 1) respect for the morphogenetic or syntactic structures of the target language, which means that the derivations or the structures used in translation must be adjusted to the traditional patterns in TL; 2) respect for the pre-existing vocabulary in the TL, which means rejecting the creation of ' unnecessary' terms, and the need to coin neologisms derived from pre-existing TL terms; 3) respect for the semantic logic of the resulting term as compared with similar terms already existing in TL. (Alvarez Lugrís, 199: P. 188)

Normally, learned prescriptivists possess the ability to furnish reasons that do not seem arbitrary, purely aestheticist, or disproportionately nationalist. On the contrary, their arguments are full of common sense and based on a sound knowledge of TL dynamics, so it is difficult not to admit at least that things would be much more orderly and logical if they were the way they should be according to the prescriptivists' view. The example of ' randomizado/aleatorizado' for the English ' randomis ed' mentioned above should illustrate this clearly. In Spanish, the scientific sense of ' random' is usually translated as ' aleatorio' , a quite common term in statistics. Thus, it makes no sense to coin the (very opaque for outsiders) neologism ' randomizado' when it would be very easy to extend the pre-existing word to ' aleatorizado'.

On the other hand, descriptivists declare that when dealing with communication the key issue is not being philologically right or the way things should be. They acknowledge that the selection of a neologism is often due (no doubt unfortunately) to reasons far removed from linguistic logic and more related to power balances, either internal (for instance the convenience of creating a strong group identity derived from the creation motive I pointed out earlier) or external (the prestige motive, represented here by English as the language of science and innovation par excellence). They believe that the role of technical writers, translators included, is not mainly pedagogical but communicative, that when forced to choose between intrinsic target language correction and communicative efficacy, the latter should dominate. In this connection, the central idea is that optimal technical or scientific communication does not consist of choosing the best decontextualized terms but of ensuring the clarity and precision of the information received by the addressee. In this sense, clarity and precision mean adapting to the specialized readers' expectations and usage, instead of forcing them to guess what the 'correct' choice means or having to cope with a syntax that may easily seem inappropriate as compared with current usage in the genre, no matter how much the non-usual options might be sanctioned by tradition and TL-respect. If one accepts this view, it will often be necessary to reject the ' correct' version in favor of the one that is actually used. To illustrate this with an example of a phrase that even prescriptivists have come to accept in spite of its lack of linguistic logic according to the target language, in Spanish it has to be ciencia ficción (science fiction) instead of ficción científica (scientific fiction), which undoubtedly should be the way to construct this phrase in Spanish, where the grammar does not envisage the possibility of using nouns as adjectives. But the fact that the term came from English, possibly through French (double prestige), and that there was no pre-existing term in Spanish brought about the decision. (Ballard, 1999: PP. 42, 48)

## Development of Prescriptivism and Descriptivism in the Choice of Terminology

Nowadays 'hard' prescriptivism is not frequent, at least in Spain. The most usual purist stance now is to accept the inevitable and fight only those battles which can be won, that is, those cases in which several real terminological options make it possible to choose without jeopardizing clarity, precision and, last and least, acceptability. In Spanish medical prose, this would be, for instance, the case of droga/fármaco (drug/medicine), where droga has traditionally only been used for illegal, narcotic, and/or addictive substances, or the already explained case of randomización/aleatorización (randomization).

If 'purist' is the usual derogative term for prescriptivist translators, 'frequentist' is the one applied to descriptivists. The main potential problem of acceptability in the case of frequentism is not taking into account that usage is not unidimensional but multilayered, a source of constant headaches for translators, especially when novice - a term may be very frequent on a popular level but rejected by specialized readers as not precise or in-house enough. To choose a popular variant when translating for a specialized text basing oneself on an indiscriminate Google search, i.e. restricting oneself to the sheer amount of hits as the definitive criterion, is usually a source of problems regarding readership acceptance in technical and scientific translation because the translators will very likely find themselves terminologically off-bounds.

Professional technical and scientific translators tend to be essentially descriptivist and, at the same time, attempt to achieve a balance between what could be termed intrinsic correction from the point of view of the structure, patterns, and semantic logic of the TL, and real use from a communicative perspective. This means combining quantitative and qualitative filters when searching for terminology on the Internet or in the pertinent bibliography.

As always in translating, eclecticism ultimately rules. In practice, professional scientific and technical translators are usually aware that there tends not to be a unique terminological solution, and that there are at least two very different perspectives on the issue of interference. Normally, a professional knows that if you make sure that your term is really used or, even better, preferred in the text-type domain you are translating for (thus guaranteeing clarity and acceptability) and that it really means what you want it to convey (ensuring precision), things should work fine. If, on top of that, you are able to choose the most logical and TL-respectful alternative because it is in fact used and accepted by your readership, so much the better. Of course, on many occasions, this is much easier said than done. It should be possible to combine quality and quantity filters in documentation, but one must be prepared to receive criticism from both poles since there will almost always be an alternative preferred by the other side. (Gottlieb, 2001: PP. 249, 258)

## Discussion

Principle interference in technical and scientific translation is one of the crucial findings in the recent researches in the sphere of translation as Jerome says that in the Bible even the order of the words is sacred and should be respected. Schleiermacher, another theologian, is probably the first scholar to defend in a systematic way what could be termed 'controlled interference' in the translation of scientific, technical, canonical and sacred texts for the same reasons. He was also probably the first to explicitly exclude technical texts from this kind of strategy since their wording supposedly did not convey any special national spirit and their translation Swas a mainly mechanical task.

In modern times, scholars such as Benjamin, Berman, and Venuti have retaken Schleiermacher's stance from different starting points and ideological agendas to favor overt translations enabling the reader to perceive the source text as portraying a different culture. These authors denounce normalization (i.e. the replacement of foreign or idiosyncratic marks included in the source text by the most usual variants according to target text conventions) as a strategy that eliminates "otherness" from a foreign text which should also convey a different worldview for TL addressees. Normalization, then, would result in target texts all written in a uniform way, giving the impression that all literature and views of life are essentially the same. While I believe these limitations have not impacted the primary outcome of the study, future work could seek to include additional interference in scientific and technical translation.

## Conclusion

Interference in technical and scientific translation has always been a topic of great interest in the theory of translation, although considered from different perspectives and under different labels, some of them even more value-laden than "interference" itself. Lexical and syntactic interference in particular has traditionally been regarded as classic howlers, something to be systematically avoided because it worked against a fluent and transparent reading.

Technology unexpectedly has developed and connected people around the world. Therefore, translation is the most important tool in order to solve communicative challenges and difficulties among people around the world. Hence, technical translation is one of the most important parts of the proper transition to escape from ambiguity and it helps to understand the opponent clearly. Many updated sciences are available in different languages and the only tool that can help us to benefit from that information in translation. On the other hand, powerful societies and governments are working to influence other communities.

Normally, learned prescriptivists possess the ability to furnish reasons that do not seem arbitrary, purely aestheticist, or disproportionately nationalist. They are donating fully-funded scholarships and other plans to develop their language, tradition, culture, and other social activities. These activities can be done with the help of translation and the term of technical and scientific translation methods are essential parts of translation where it can help the language learners to understand the scope properly and authentically. I have just compiled about available problems and difficulties of technical translation. There are scientific methods to render those words properly and understandably from the source language to the target language. It is really necessary to learn those who are working as a translator or interpreters because without those methods it is too hard to render a word correctly to the target language.